\documentclass[runningheads,a4paper]{llncs}

\usepackage{amssymb}
\usepackage{graphicx}
\usepackage{url}

\usepackage{amsmath}
\usepackage{algorithmic}
\usepackage{algorithm}
\usepackage{subfigure}
\usepackage{array}
\usepackage{tabularx}
\usepackage{multirow}

\newcommand{\bfx}{{\textbf{x}}}
\newcommand{\bfv}{{\textbf{v}}}

\newcommand{\bft}{{\textbf{t}}}

\newcommand{\keywords}[1]{\par\addvspace\baselineskip
\noindent\keywordname\enspace\ignorespaces#1}

\begin{document}

\mainmatter  % start of an individual contribution

\title{Image tag completion by local learning}

\titlerunning{Image tag completion by local learning}

\author{Jingyan Wang$^{1,2,3}$
\and
Yihua Zhou$^4$
\and
Haoxiang Wang$^5$
\and
Xiaohong Yang$^6$
\and
Feng Yang$^6$
\and
Austin Peterson$^7$
}

\authorrunning{J. Wang, et al.}

\institute{
National Time Service Center, Chinese Academy of Sciences, Xi' an, Shaanxi 710600 , China\\
\email{jingbinwang1@outlook.com}
\and
Graduate University of Chinese Academy of Sciences, Beijing 100049, China
\and
Provincial Key Laboratory for Computer Information Processing Technology, Soochow University Suzhou 215006, China
\and
Department of mechanical engineering and mechanics, Lehigh University, Bethlehem, PA 18015, US
\and
Department of Electrical and Computer Engineering, Cornell University, Ithaca, NY 14850, USA
\and
College of  Computer Science and Technology, Shandong University of Finance and Economics, Jinan 250014, China
\and
Electrical and Computer Engineering Department, The University of Texas at San Antonio, San Antonio, TX, 78249, USA\\
\email{austin.peterson1@outlook.com}
}

%\toctitle{Lecture Notes in Computer Science}
%\tocauthor{Authors' Instructions}
\maketitle

\begin{abstract}
The problem of tag completion is to learn the missing tags of an image. In this paper, we propose to learn a tag scoring vector for each image by local linear learning. A local linear function is used in the neighborhood of each image to predict the tag scoring vectors of its neighboring images. We construct a unified objective function for the learning of both tag scoring vectors and local linear function parameters. In the objective, we impose the learned tag scoring vectors to be consistent with the known associations to the tags of each image, and also minimize the prediction error of each local linear function, while reducing the complexity of each local function. The objective function is optimized by an alternate optimization strategy and gradient descent methods in an iterative algorithm. We compare the proposed algorithm against different state-of-the-art tag completion methods, and the results show its advantages.
\end{abstract}

\section{Introduction}

Recently, social network has been a popular tool to share images. When a social network user uploads an image, the image is usually associated with a tag/ keyword which is used to describe the semantic content of this image. The tags provided by the users are usually incomplete. Zhang et al. designed and implemented a fast motion detection mechanism for multimedia data on mobile and embedded environment \cite{zhang2014lucas}. Recently, the problem image tag completion is proposed in the computer vision and machine learning communities to learn the missing tags of images \cite{Wu2013716,Lin20131618,Lin201442,Feng2014424,Xia2014,liu2015supervised,wang2015representing}. This problem is defined as the problem of complete the missing elements of a tag vector of a given image automatically.

In this paper, we investigate the problem of image tag completion, and proposed a novel and effective algorithm for this problem based on local linear learning.
We propose a novel and effective tag completion method. Instead of completing the missing tag association elements of each image, we introduce a tag scoring vector to indicate the scores of assigning the image to the tags in a given tag set. We propose to study the tag scoring vector learning problem in the neighborhood of each image. For each image in the neighborhood, we propose to learn a linear function to predict a tag scoring vector from a visual feature vector of its corresponding image feature. We propose to minimize the perdition error measure by the squared $\ell_2$ norm distance over each neighborhood, and also minimize the squared $\ell_2$ norm of the linear function parameters. Besides the local linear learning, we also proposed to regularize the learning of tag scoring vectors by the available tags of each image. We construct a unified objective function to learn both the tag scoring vectors and the  local linear functions. We develop an iterative algorithm to optimize the proposed problem. In each iteration of this algorithm, we update the tag scoring vectors and the local linear function parameters alternately.

This rest parts of  paper are organized as follows: in section \ref{sec:method}, we introduced the proposed method. In section \ref{sec:exp}, we evaluate the proposed methods on some benchmark data sets. In section \ref{sec:conclusion}, the paper is concluded with future works.

\section{Proposed method}
\label{sec:method}

We assume that we have a data set of $n$ images, and their visual feature vectors are $\bfx_i|_{i=1}^n$, where $\bfx_i \in \mathbb{R}^d$ is the $d$-dimensional feature vector of the $i$-th image. We also assume that we have a set of $m$ unique tags, and a tag vector $\widehat{\bft}_i =[\widehat{t}_{i1},\cdots, \widehat{t}_{im}]^\top \in \{+1,-1\}^m$ for the $i$-th image $\bfx_i$, where
$\widehat{t}_{ij} =
+1$ if~the~$j$-th~tag~is~assigned ~to~the~$i$-th~image, and $-1$, otherwise. In real-world applications, the tag vector of an image $\bfx_i$ is usually incomplete, i.e., some elements of $\widehat{\bft}_i$ are missing. We define a vector $\bfv_i = [v_{i1},\cdots,v_{im}]\in \{1,0\}^m$, where $v_{ij} = 1$ if $\widehat{t}_{ij}$ is available, and $0$ if $\widehat{t}_{ij}$ is missing. We propose to learn a tag scoring vector $\bft_i = [t_{i1}, \cdots,t_{im}]\in \mathbb{R}^m$, where $t_{ij}$ is the score of assigning the $j$-th tag to the $i$-th image.

The set of the $\kappa$ nearest neighbor of each image $\bfx_i$ is denoted as $\mathcal{N}_i$, and we assume that the tag scoring vector $\bft_j$ of a image $\bfx_j \in \mathcal{N}_i$ can be predicted from its visual feature vector $\bfx_j$ using a local linear function $f_i(\bfx_j)$,

\begin{equation}
\begin{aligned}
\bft_j  \leftarrow f_i(\bfx_j)=W_i \bfx_j , ~\forall~j:\bfx_j\in \mathcal{N}_i,
\end{aligned}
\end{equation}
where $W_i\in \mathbb{R}^{m\times d}$ is the parameter of the local linear function. To learn the tag scoring vector and the local function parameters, we propose the following minimization problem,

\begin{equation}
\label{equ:obj3}
\begin{aligned}
\min_{\bft_i|_{i=1}^n,W_i|_{i=1}^n}
&
\left \{
g(\bft_i|_{i=1}^n,W_i|_{i=1}^n)
\vphantom{\sum_{j:v_{ij}=1}}
 =
\sum_{i=1}^n
\left (
\sum_{j:\bfx_j\in \mathcal{N}_i} \| \bft_j - W_i \bfx_j\|_2^2 + \alpha \|W_i\|_2^2
\right.\right.
\\
&\left .\left .
\vphantom{\sum_{i}^2}
+\beta(\bft_i - \widehat{\bft}_i)^\top diag(\bfv_i) (\bft_i - \widehat{\bft}_i)
\right )
\right \}
\end{aligned}
\end{equation}
where $\alpha$ and $\beta$ are tradeoff parameters. The objective function $g(\bft_i|_{i=1}^n,W_i|_{i=1}^n)$ in (\ref{equ:obj3}) is a summarization of three terms over all the images in the data set. The first term, $\sum_{j:\bfx_j\in \mathcal{N}_i} \| \bft_j - W_i \bfx_j\|_2^2$, is the prediction error term of the local linear predictor over the neighborhood of each image. The second, $\|W_i\|_2^2$, is to reduce the complexity of the local linear predictor. The last term, $(\bft_i - \widehat{\bft}_i)^\top diag(\bfv_i) (\bft_i - \widehat{\bft}_i)$, is a regularization term to regularize the learning of tag scoring vectors by the incomplete tag vectors, so that the available tags are respected.
To optimize the minimization problem in (\ref{equ:obj3}), we propose to use the alternate optimization strategy \cite{Huang2015233,Liu2015188} in an iterative algorithm.

\begin{itemize}
\item \textbf{Optimization of $\bft_i|_{i=1}^n$}
In each iteration, we optimize $\bft_i|_{i=1}^n$ one by one, and the minimization of (\ref{equ:obj3}) with respect to $\bft_i$ can be achieved with the following gradient descent update rule,

\begin{equation}
\label{equ:ti}
\begin{aligned}
\bft_i^{new} = \bft_i^{old} - \eta \nabla_{\bft_i}g
(\bft_j|_{j=1}^n,W_i|_{i=1}^n)
|_{\bft_i=\bft_i^{old}},
\end{aligned}
\end{equation}
where $\nabla_{\bft_i}g(\bft_j|_{j=1}^n,W_i|_{i=1}^n)$ is the sub-gradient function of $g(\bft_j|_{j=1}^n,W_i|_{i=1}^n)$, with respect to $\bft_i$,

\begin{equation}
\label{equ:gradient}
\begin{aligned}
\nabla_{\bft_i}g
(\bft_j|_{j=1}^n,W_i|_{i=1}^n)
= 2 \sum_{k:\bfx_i \in \mathcal{N}_k} (\bft_i - W_k \bfx_i) + 2\beta diag(\bfv_i) (\bft_i - \widehat{\bft}_i),
\end{aligned}
\end{equation}
and $\eta$ is the descent step.

\item \textbf{Optimization of $W_i|_{i=1}^n$}
In each iteration, we also optimized $W_i|_{i=1}^n$ one by one. When $W_i$ is optimized, $W_j|_{j\neq i}$ are fixed. Gradient descent method is also employed to update $W_i$ to minimize the objective in (\ref{equ:obj3}),

\begin{equation}
\label{equ:w1}
\begin{aligned}
W_i^{new} = W_i^{old} - \eta \nabla_{W_i} g(\bft_i|_{i=1}^n, W_i|_{i=1}^n)|_{W_i=W_i^{old}},
\end{aligned}
\end{equation}
where $\nabla_{W_i} g(\bft_i|_{i=1}^n, W_i|_{i=1}^n)$ is the sub-gradient function with respect to $W_i$,

\begin{equation}
\label{equ:w2}
\begin{aligned}
\nabla_{W_i} g(\bft_i|_{i=1}^n, W_i|_{i=1}^n)
= 2 \sum_{j:\bfx_j \in \mathcal{N}_i} (\bft_j - W_i \bfx_j) \bfx_j^\top + 2 \beta W_i.
\end{aligned}
\end{equation}

\end{itemize}

\section{Experiments}
\label{sec:exp}

\subsection{Setup}

In the experiments, we used two publicly accessed image-tag data sets, which are Corel5k data set \cite{Zhang2012838,Huang20103376,Wang20091643} and IAPR TC12 data set \cite{Zhang20151658,Li20132700,Zhang2012838}.
In the Corel5k data set, there are 4,918 images, and 260 tages. We extract density feature, Harris shift feature, Harris Hue feature, RGB color feature, and HSV color feature as visual features for each image. Moreover, we remove  40\% of the elements of the tag vectors to make the incomplete image tag vectors. In the IAPR TC12 data set, there are 19,062 images, and 291 tags. We also remove 40\% elements of the tag elements to construct the incomplete tag vectors.
To evaluate the tag completion performances, we used the recall-precision curve as performance measure.
We also use mean average precision (MAP) as a single performance measure.

\subsection{Results}

\begin{figure}[!htb]
\centering
\subfigure[Recall-precision curve]{
\includegraphics[width=0.45\textwidth]{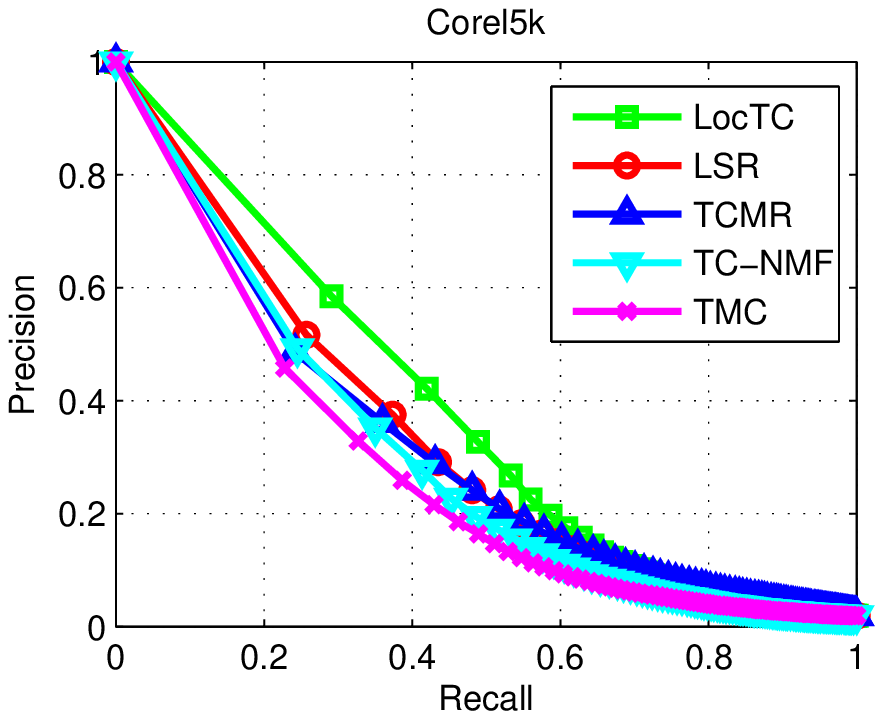}}
\subfigure[MAP]{
\includegraphics[width=0.45\textwidth]{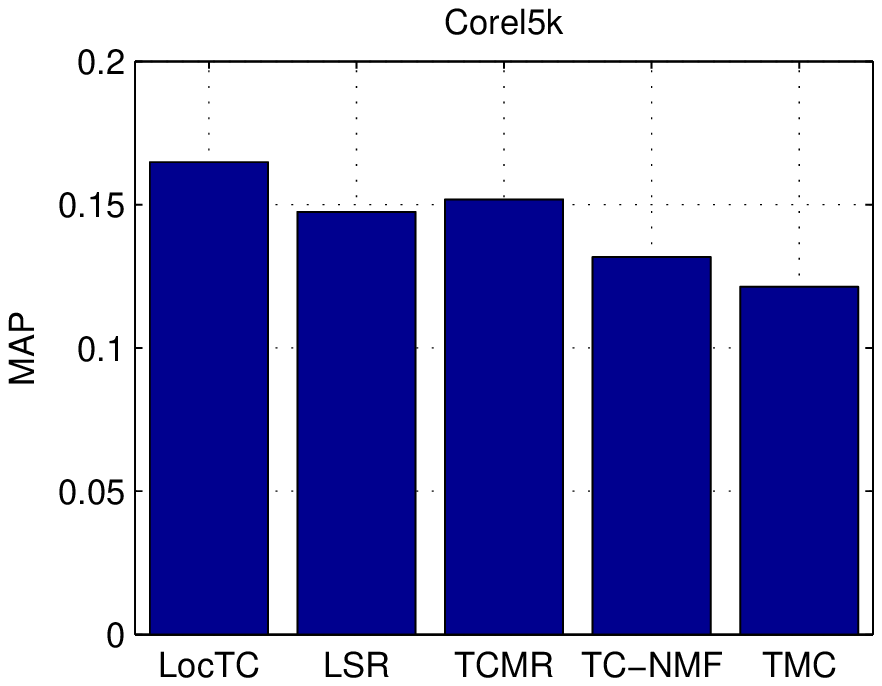}}
\caption{Results of comparison to state-of-the-art methods on Corel5k data set}
\label{fig:Corel5k}
\end{figure}

\begin{figure}[!htb]
\centering
\subfigure[Recall-precision curve]{
\includegraphics[width=0.45\textwidth]{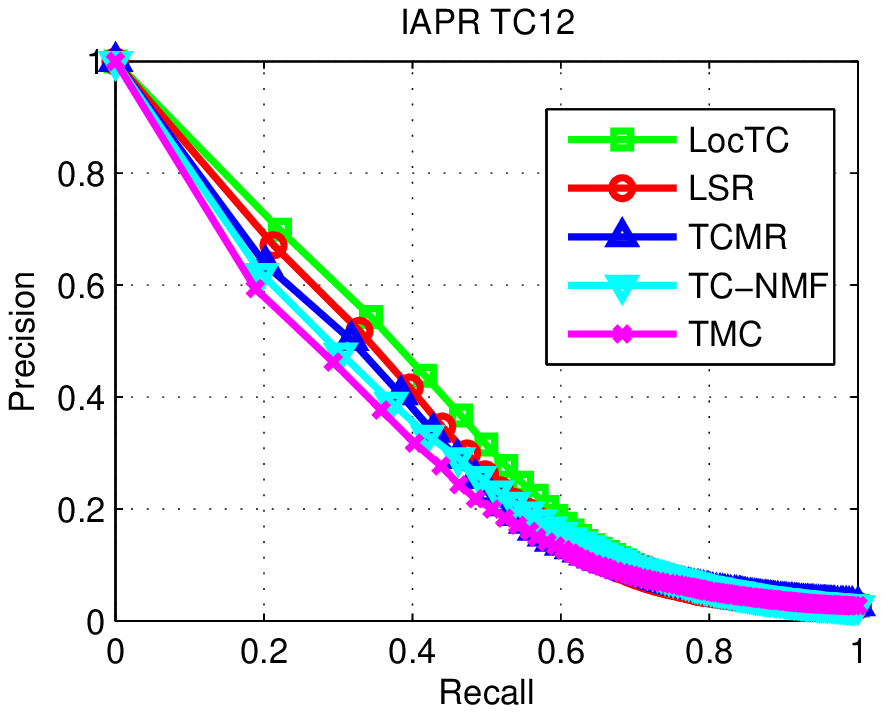}}
\subfigure[MAP]{
\includegraphics[width=0.45\textwidth]{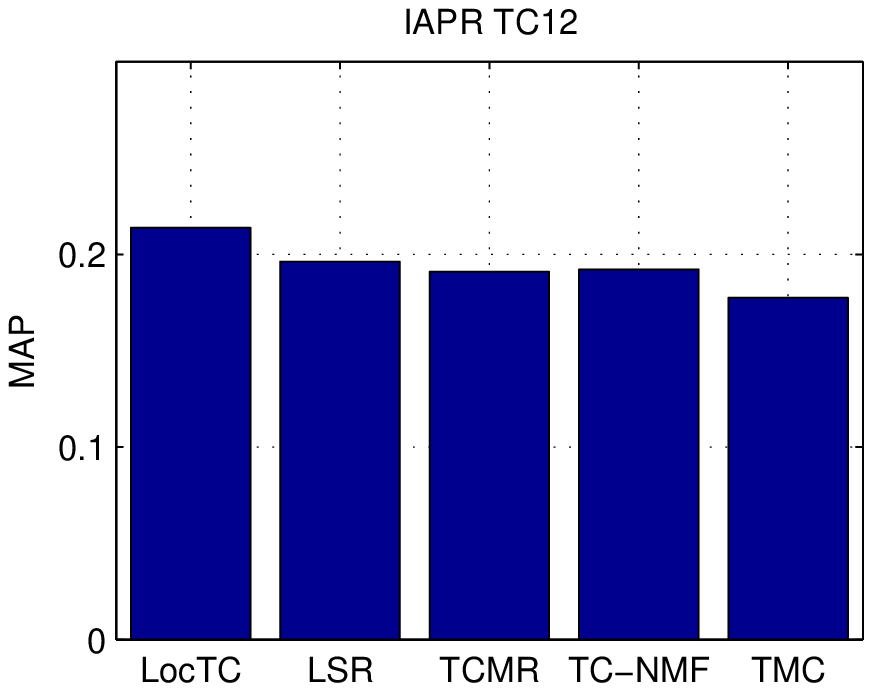}}
\caption{Results of comparison to state-of-the-art methods on IAPR TC12 data set}
\label{fig:IAPR}
\end{figure}

We compared the proposed method to several state-of-the-art tag completion methods, including tag matrix completion (TMC) \cite{Wu2013716}, linear sparse reconstructions (LSR) \cite{Lin201442}, tag completion by noisy matrix recovery (TCMR)\cite{Feng2014424}, and tag completion via NMF (TC-NMF) \cite{Xia2014}. The experimental result on two data sets are given in Fig. \ref{fig:Corel5k} and Fig. \ref{fig:IAPR}. From these figures, we can see that the proposed method LocTC performs best. Its recall-precision curve is closer to the top-right corner than any other methods, and its MAP is also higher than MAPs of other methods.

\begin{figure}[!htb]
  \centering
\subfigure[$\alpha$]{
\includegraphics[width=0.45\textwidth]{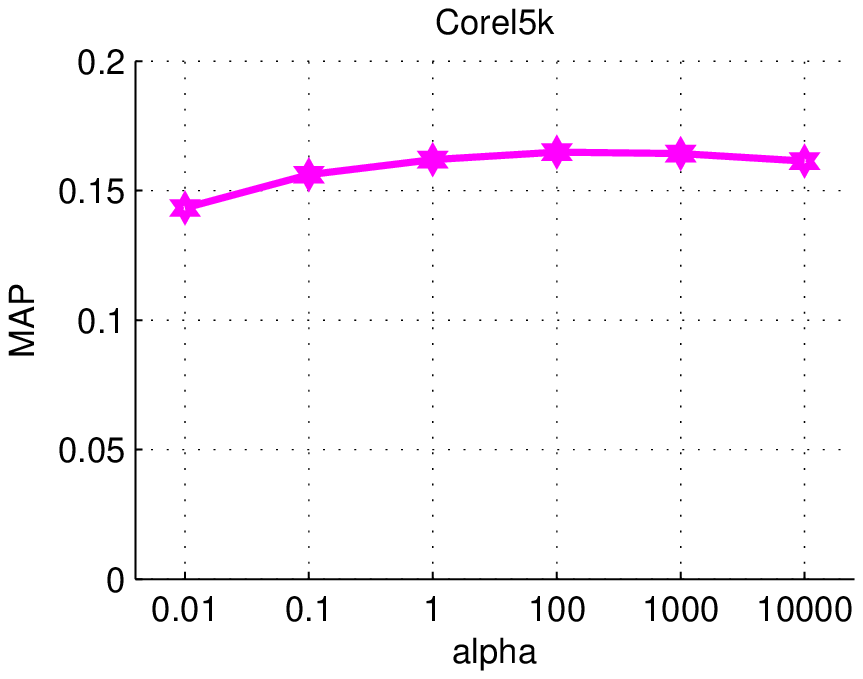}}
\subfigure[$\beta$]{
\includegraphics[width=0.45\textwidth]{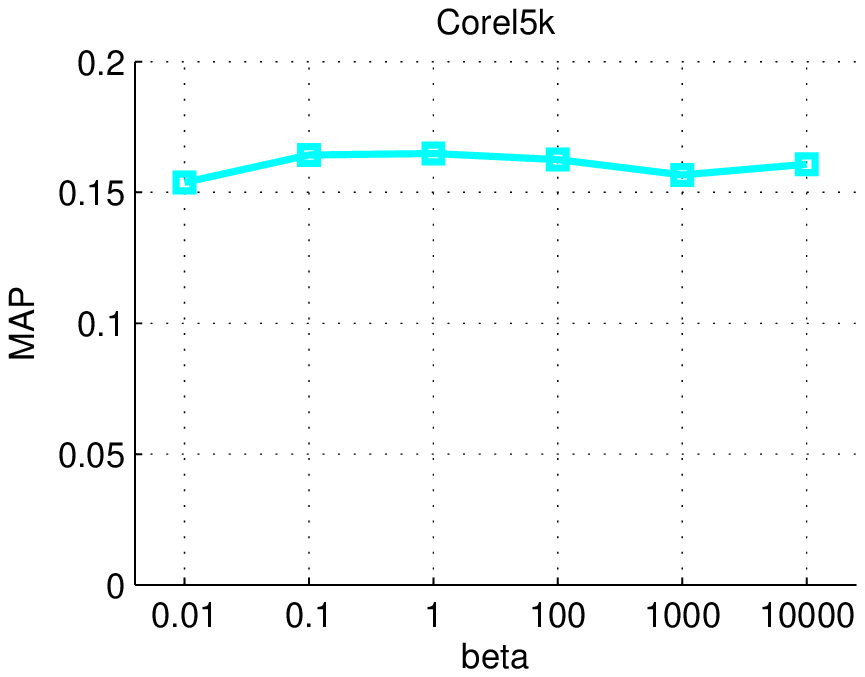}}
\\
\caption{Parameter sensitivity curve on Corel5k data set.}
\label{fig:Corel_alpah}
\end{figure}

\begin{figure}[!htb]
  \centering
\subfigure[$\alpha$]{
\includegraphics[width=0.45\textwidth]{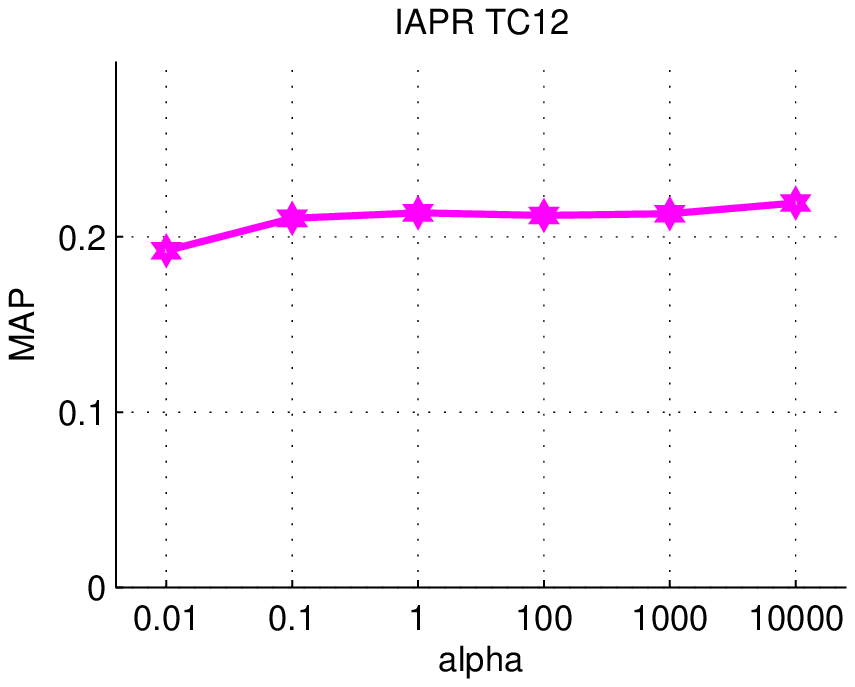}}
\subfigure[$\beta$]{
\includegraphics[width=0.45\textwidth]{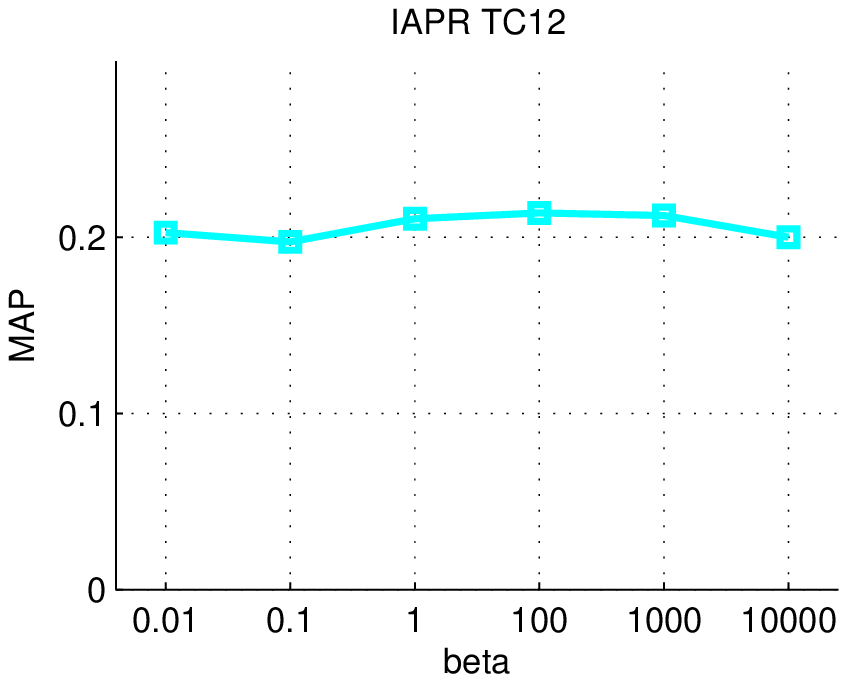}}
\\
\caption{Parameter sensitivity curve on IAPR TC12 data set.}
\label{fig:IAPR_alpah}
\end{figure}

In this section, we will study the sensitivity of the proposed algorithm to the two parameters, $\alpha$ and $\beta$. The curves of $\alpha$ and $\beta$ on different data sets are given in Fig. \ref{fig:Corel_alpah} and Fig. \ref{fig:IAPR_alpah}. From these figures, we can see that the performances are stable to different valuse of  both $\alpha$ and $\beta$.

\section{Conclusion and future works}
\label{sec:conclusion}

In this paper, we study the problem of tag completion, and proposed a novel algorithm for this problem. We proposed to learn the tags of images in the neighborhood of each image. A local linear function is designed to predict the tag scoring vectors of images in each neighborhood, and the prediction function parameter is learned jointly with the tag scoring vectors. The proposed method is compared to state-of-the-art tag completion algorithms, and the results show that the proposed algorithm outperforms the compared methods. In the future, we will study how to incorporate these connections into our model and learn more effective tags. In this paper, we used one single local function for each neighborhood, and in the future, we will use more than than regularization to regularized the learning of tags \cite{wang2012multiplegraph,wang2013multiple}, such as usage of wavelet functions to construct the local function \cite{liu2012wavpeak}. Moreover, correntropy can also be considered as a alternative loss function to construct the local learning problem \cite{wang2013non,Xing2014483,Li2015850,Zhang2015120,wang2014effective}. In the future, we also plan to extend the proposed algorithm for completion of tags of large scale image data set by using high performance computing technology \cite{zhou2013exploring,wang2015towards,zhang2011gpapriori,zhang2013accelerating,zhang2011frequent,zhang2013fpga,gao2014sparse,zhang2014lucas,li2013zht,zhao2014fusionfs,li2013distributed,wang2013using,wang2014optimizing,wang2014next}, and completion of tags of gene/protein functions of bioinformatics problems \cite{hu2009improving,zhang2009bayesian,zhang2010bioinformatics,hu2009improving,wang2015multiple,wang2015supervised}.

%\bibliographystyle{spmpsci}
%\bibliography{Tag150106}

\end{document}